\documentclass{article}
\PassOptionsToPackage{numbers,sort&compress}{natbib}

\usepackage[preprint]{nips_2018}

\usepackage[utf8]{inputenc} 
\usepackage[T1]{fontenc}    
\usepackage{hyperref}       
\usepackage{url}            
\usepackage{booktabs}       
\usepackage{amsfonts}       
\usepackage{nicefrac}       
\usepackage{microtype}      
\usepackage[pdftex]{graphicx}
\usepackage{subfig}
\usepackage{amsmath}
\usepackage{authblk}
\usepackage{enumitem}

\graphicspath{{./}}

\title{Generalizing multistain immunohistochemistry tissue segmentation using one-shot color deconvolution deep neural networks}

\author[1,3]{Amal Lahiani}
\author[2]{Jacob Gildenblat}
\author[1]{Irina Klaman}
\author[3]{Nassir Navab}
\author[1]{Eldad Klaiman}
\affil[1]{Pathology and Tissue Analytics, Pharma Research and Early Development, Roche Innovation Center Munich}
\affil[2]{SagivTech LTD., Israel}
\affil[3]{Computer Aided Medical Procedures, Technische Universit\"at M\"unchen, Germany}
\affil[ ]{\textit {\{amal.lahiani, irina.klaman, eldad.klaiman\}@roche.com, jacob@sagivtech.com, navab@cs.tum.edu}}

\begin{document}
\maketitle

\footnote{This work has been submitted to the IEEE for possible publication. Copyright may be transferred without notice, after which this version may no longer be accessible.}
\begin{abstract}
  A key challenge in cancer immunotherapy biomarker research is quantification of pattern changes in microscopic whole slide images of tumor biopsies. Different cell types tend to migrate into various tissue compartments and form variable distribution patterns. Drug development requires correlative analysis of various biomarkers in and between the tissue compartments. To enable that, tissue slides are manually annotated by expert pathologists. Manual annotation of tissue slides is a labor intensive, tedious and error-prone task. Automation of this annotation process can improve accuracy and consistency while reducing workload and cost in a way that will positively influence drug development efforts. In this paper we present a novel one-shot color deconvolution deep learning method to automatically segment and annotate digitized slide images with multiple stainings into compartments of tumor, healthy tissue, and necrosis. We address the task in the context of drug development where multiple stains, tissue and tumor types exist and look into solutions for generalizations over these image populations.
\end{abstract}

\section{Introduction}

Recent studies in digital pathology have shown significant advancement in automatic tumor detection in clinical diagnostic scenarios (\citet{21, 9}). However, the unique requirements for research and drug development are not addressed in most of these studies. This is because typically, in a clinical diagnostic environment, the tasks are limited to a single staining, typically Hematoxylin and Eosin (H\&E) stained slides, from a specific hospital lab and containing one kind of tissue with a specific cancer type (e.g. Metastases in breast) (\citet{21, 9}). In the context of drug development, and specifically in the development of immunotherapy drugs, the samples are much more variable. Slides are typically stained in at least 3 different Immunohistochemistry (IHC) staining methods with different stain types and colors in addition to the traditional H\&E. Moreover, the slides originate from different hospital centers around the globe which can lead to variability in the fixation and state of the specimens due to varying collection techniques. It is also common to have 5-6 cancer types in a clinical trial and biopsies or surgical specimen may come from various body organs. Manual annotation of tissue slides is a time consuming and poorly-reproducible task, especially when dealing with images scanned under high magnification and containing billions of pixels. Previous studies have shown that pathologist concordance in the manual analysis of slides suffers from a relatively high inter and intra user variability (\citet{16, 3,20}) mainly due to the large size of the images. Automation of this annotation process can improve the accuracy and consistency while reducing workload and cost in a way that will positively influence drug development efforts and will assist pathologists in the time consuming operation of manual segmentation. In the recent years, deep learning methods have been used with great success in several domains, including computer vision (\citet{22, 23,24}), sometimes surpassing human capabilities on tasks such as classification performed on specific data sets. Motivated by the success of applying deep learning based approaches to image segmentation tasks (\citet{2}), we investigate the possibility of applying deep learning based algorithms for automatic tissue segmentation.

In the next sections of this paper, we describe the main building blocks used for our automatic tissue segmentation algorithm including a deep learning network for segmentation of a highly unbalanced dataset with 4 categories, a methodology for addressing the dataset and its high variability (9 stain types were included in our experiments), a one-shot color deconvolution deep learning network architecture and visualization techniques for understanding the algorithm results and gaining trust from pathologists to use the system.

\section{Dataset}
We selected 77 whole slide images of Colorectal Carcinoma metastases in liver tissue from biopsy slides stained with H\&E (blue,pink) and 8 additional immunohistochemistry (IHC) assays stains as follows: CEA (brown), CD163/CD68 (brown, red), CD8/CD3 (brown, red), FoxP3 (brown), Ki67/CD3 (brown, red), Ki67/CD8 (purple, yellow), PRF/CD3 (brown, red), PD1 (brown). All these IHC stainings use a blue (Hematoxylin) counterstain. The selection was done according to the most frequent stainings. These 77 slides compose our dataset for this project. We then split this dataset into training (51 slides) and testing (26  slides) sets. The various regions on the slides were annotated with one of the following categories: "Tissue" - i.e. normal tissue, "Tumor", "Necrosis" - i.e. dead  tissue, and "Exclude" - i.e. areas not to be used due to irrelevance, artifacts, etc. Areas not included in any of the above categories is labeled "Background".

\begin{figure}
\centering
\includegraphics[width=0.98\linewidth]{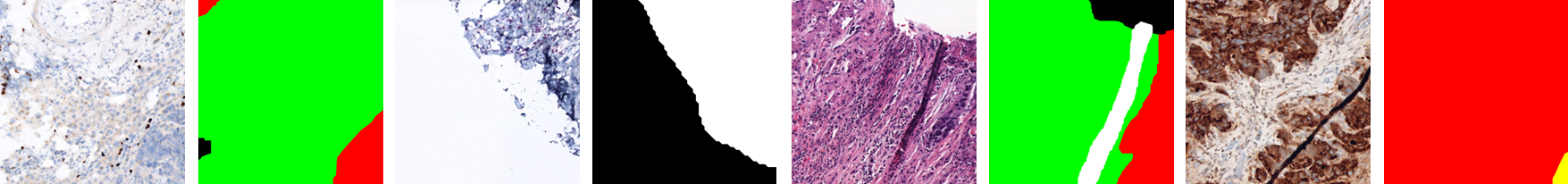}
\caption{Example tiles from the dataset and their respective label images colored green, red, yellow, black and white corresponding to the different tissue regions of tissue, tumor, necrosis and background and exclude respectively.   
}
\label{image_label}
\end{figure}

Each high resolution whole slide image was split into overlapping $512\times512\times3$ RGB tiles for processing at a 10x magnification factor (half the original scanning resolution). The selection of a lower resolution was needed to increase the contextual information when classifying a given tissue pixel. The memory bound on the compute hardware limits the size of the input images to the network and therefore, in order to have enough tissue context for a limited tile size in the image dataset, we opted for the reduced magnification. Other studies have also found 10x magnification to be sufficient for tissue segmentation tasks (\citet{21, 9}). For each image tile a corresponding ground truth tile was created using the region annotations of the pathologist. This process yielded 16834 $512\times512\times3$ RGB tiles and their corresponding  $512\times512\times1$ ground truth tiles. A sample of these tiles can be seen in figure \ref{image_label}.

Different biopsies contain variable amounts of background, tumor, tissue and necrosis. Thus, the generated dataset contained an inherent imbalance of the classes (58\%, 19\%, 16\% and 7\% for background, tumor, tissue and necrosis respectively). Such class imbalances can significantly hinder the learning process in deep learning. To address the high class imbalance we use a loss weighting strategy called median frequency balancing (\citet{4}). 

In order to increase the amount of labeled data for training the network, we used data augmentation techniques including hue, brightness and scale jittering, adding uniform noise to the intensity channel, random occlusion, image flips and random but small affine transforms.

\section{Methodology and results}

This section describes our methodology, the neural network used for the tissue segmentation task and the results we obtained.
The input images are BGR $512\times512$ images scaled to the range [0, 1], and the outputs are $512\times512\times4$ images, where each pixel contains a vector of four probabilities for that pixel for each class.  
The loss function is the average of the cross entropy loss, across the image at the output from the network. The loss of pixels marked to be ignored, were given a zero weight. The optimization step was done with the SGD optimizer (\citet{25}) with a momentum of 0.9. 
To accelerate the network training on multiple GPUs, the Synchronous SGD algorithm (\citet{26}) was used, using 15 GPUs on the Roche Pharma HPC cluster in Penzberg and a total batch size of 240 (16 per GPU). The base learning rate of the SGD optimizer was multiplied  by the number of the GPUs used, where 0.001 was the default learning rate for 1 GPU. We used the distributed module in PyTorch (\citet{1}) for implementing the distributed training, which was crucial for fast experimenting on large quantities of data. During training, 10\% of the training data was reserved for validation. As an evaluation metric we used the F1 score to account for both precision and recall. 

\subsection{UNET with a single staining vs UNET with multiple stainings}
\begin{figure}
\centering
\includegraphics[width=0.6\linewidth]{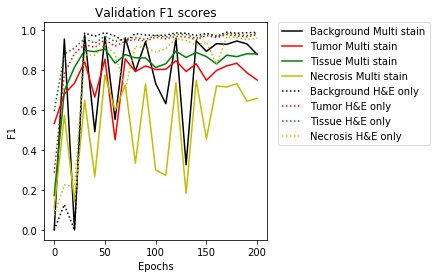}
\caption{Evolution of  validation F1 scores during training. The black, red, green and yellow curves correspond to background, tumor, tissue and necrosis respectively. The dotted and solid lines correspond to validation scores of UNET during training for a single stain (H\&E) and a multistain dataset respectively.}
\label{hne_multi}
\end{figure}

In the development of the network architecture we initially examined the results of a segmentation network from previous research: UNET fully convolutional network (\citet{13}). We trained UNET with 2 different datasets: The first dataset contains only one specific staining (H\&E) and the second dataset contains different staining types (H\&E and 8 IHCs). We noticed that validation F1 scores of the different classes converge smoothly during training after 200 epochs in the case of the single stain dataset while they did not converge smoothly or to the same levels during the same training period in the case of the  multiple staining dataset. This difference is most strongly expressed for the classes of necrosis and background. Figure \ref{hne_multi} shows validation F1 scores for the different classes during the 200 epochs of training with both datasets. The validation and the calculation of the scores were performed every 10 epochs.

Evaluation of the networks' performance on the testing dataset showed increased generalization capabilities for the single stain dataset compared to the multistain dataset (Table \ref{F1_hne_multi}). For a fair comparison, we evaluated both networks on exactly the same testing set image using only H\&E images from the testing set.

\begin{table}
\caption{Testing F1 scores for each of the categories: multiple stain and H\&E only}
\label{F1_hne_multi}
\centering 
\begin{tabular}{lllll}
\toprule
 & Background & Tumor & Tissue & Necrosis\\ \midrule
Multistain & 0.63 & 0.20 & \textbf{0.78} & 0.44\\ \midrule
H\&E only & \textbf{0.99} & \textbf{0.85} & 0.71 & \textbf{0.88} \\ \bottomrule

\end{tabular}
\end{table}

\subsection{Network architecture: CD-UNET}
These results made us theorize that images of slides with multiple stainings in the training dataset make it harder to learn the correct features of the different classes due to increased dataset and intra-class variability. Additionally, when multiple stainings are available, it is preferred to train a single model that will perform well on all stainings rather than multiple models (i.e. for each specific stain) in order to avoid the need for storing multiple models. A combined dataset with multiple stainings also has the advantage of making the training dataset larger and more comprehensive thus potentially also reducing overfitting. Still, the slower convergence during training and the lower generalization present challenges that need to be dealt with.

One way of limiting the detrimental effect of stain variability on training which has been described in previous studies is using the concentration of stains as network input instead of the RGB pixel values. Color deconvolution has been used on histology images as a pre-processing step showing improved segmentation and classification results (\citet{27, 28, 29, 35}). However, stain deconvolution is generally subjective and depends on stain bases. In the context of drug development, staining procedures, colors and quality may vary for different reasons such as lab processes, imaging scanners and staining protocols. In \citet{36} a stain separation layer was added to the network as part of a binary cell classification framework, however a pre-processing step to estimate the optical density from the raw image was used. In addition, their method is heavily dependent on the parameter initialisation of the stain separation layer. This filter initialisation was based on stain parameters making it subjective and limited to the use of a single staining. In our case the training dataset comprises multiple stainings making the approach problematic to use. 

We propose addressing these problems by the addition of an inherent color deconvolution segment to the UNET architecture. In our approach, the color deconvolution parameters are learned as part of the segmentation network without the need to pre-process the images, making it ideal in the case of training on a multistain dataset. We then eliminate the need to run a separate pre-processing color deconvolution step for each different stain type and make both training and inference simpler. To the best of our knowledge, this is the first work where a color deconvolution estimation, applicable to different stainings, was learned inherently as part of a segmentation network without the need to pre-define the stain basis parameters. 

The proposed architecture has two additional ($1\times1$) convolution layers preceding the original UNET segmentation network. As the training dataset is composed of different stainings, the number of principle color shades in the training images was 6: pink, blue, purple, brown, yellow and red.  We chose then the first layer to have 6 ($1\times1$) filters, each filter corresponding to a color. The second layer contains 3 ($1\times1$) filters so as not to change the original UNET architecture input size. 

Many of state of the art stain separation methods are based on Beer-Lambert's law using the optical density space (\citet{37, 38, 39}). According to this law, the optical density is defined as $OD = \log_{10}(\frac{I_0}{I_t}) = \varepsilon LC$ where $OD$ is the optical density, $I_0$ is the intensity of the incident light, $I_t$ is the intensity of the light after passing through the specimen, $\varepsilon$ is the absorption coefficient, $C$ is the concentration of the absorbing substance, and $L$ is the thickness of the specimen. This law states then that the observed pixel intensity $I_t$ varies non-linearly with the concentration of staining. In order to allow the color deconvolution segment to learn this non-linearity of the physical model, each of the ($1\times1$) convolution layers is followed by a nonlinear function.

\begin{figure}
\centering
\includegraphics[width=0.7\linewidth]{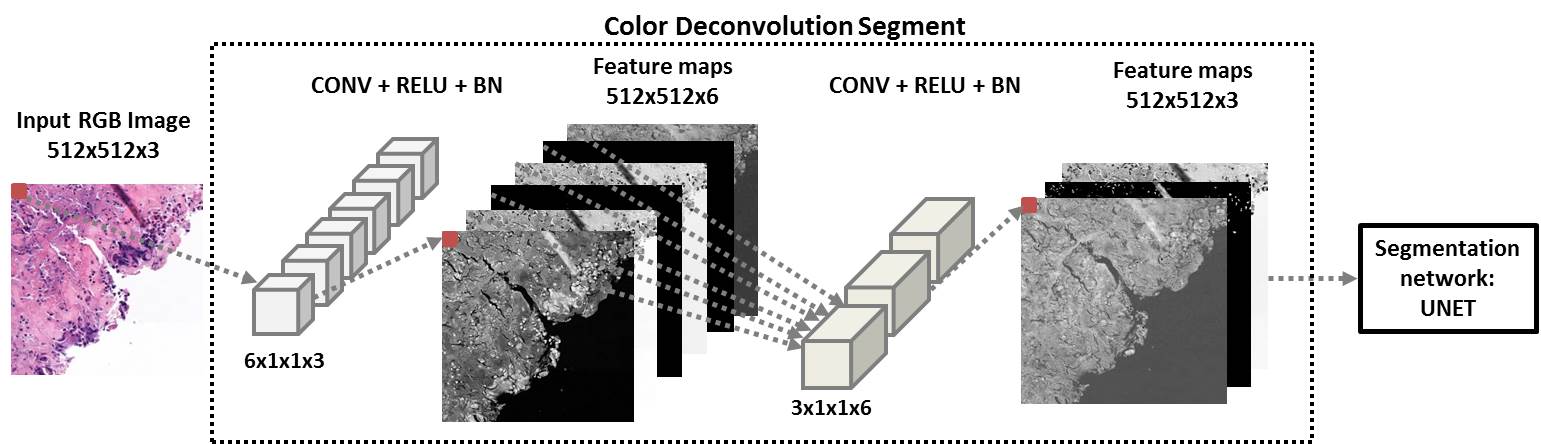}
\caption{The proposed CD-UNET architecture.}
\label{net}
\end{figure}

The proposed color deconvolution network architecture (CD-UNET) is composed of 2 main parts (Figure \ref{net}). The first part is a color deconvolution segment composed of 2 layers of ($1\times1$) convolution with ReLU and batch normalization. The second part is a UNET fully convolutional network (\citet{13})resulting in a pixel wise segmentation of the input image. We additionally modified the original UNET architecture as follows. First, we used an appropriate size of zero padding in all convolution layers to preserve the spatial size of the input to the layers. The result is that both the input and the output spatial dimensions of the network are $512\times512$. We also considered a smaller network width: each layer has half the number of filters compared to the original UNET network. This helped speeding up the learning process and reduce overfitting. In addition, batch normalization (\citet{7}) was applied after every convolutional layer.

\subsection{CD-UNET vs UNET}
\begin{figure}
\centering
\includegraphics[width=0.6\linewidth]{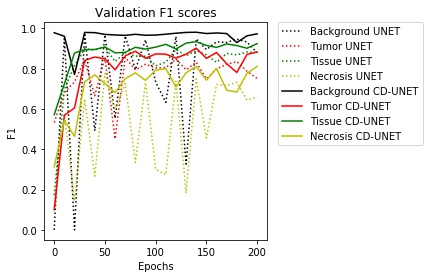}
\caption{Validation F1 scores for the different categories during 200 epochs of training. The dotted and solid lines correspond to UNET and CD-UNET respectively.}
\label{unet_cdunet}
\end{figure}

In this experiment we compare the learning and performance of UNET to CD-UNET. We trained UNET and CD-UNET for 200 epochs on the same multistain dataset and computed F1 scores for the different classes on a validation set every 10 epochs. The validation F1 scores converge faster and more smoothly on this dataset for the CD-UNET architecture compared to the original UNET architecture (Figure \ref{unet_cdunet}).

After 200 epochs of training, we evaluated the performance of both networks on a test set of 26 unseen whole slide images. The testing dataset consists of slides from the stain types that were present in the training dataset. To ensure a fair comparison, we used the same training and testing datasets for both networks. The F1 scores of the test set are listed in Table \ref{F1_unet_cdunet}. 

\begin{table}
\caption{Testing F1 scores for each of the categories : UNET and CD-UNET}
\label{F1_unet_cdunet}
\centering 
\begin{tabular}{lllll}
\toprule
 & Background & Tumor & Tissue & Necrosis\\ \midrule
UNET & 0.92 & 0.52 & 0.89 & 0.60\\ \midrule
CD-UNET & \textbf{0.99} & \textbf{0.88} & \textbf{0.90} & \textbf{0.80} \\ \bottomrule
\end{tabular}
\end{table}

F1 scores on the testing set were remarkably higher with CD-UNET and  convergence of F1 scores during training was quicker and smoother. This shows better generalization capabilities and faster learning over multistain images for the CD-UNET architecture compared with the original UNET architecture.

\subsection{Segmentation results: CD-UNET}
\begin{figure}
\centering

\subfloat[]{\includegraphics[width=.16\linewidth]{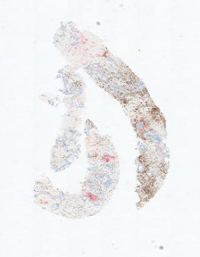}%
\label{orig_ihc}}
\hfill
\subfloat[]{\includegraphics[width=.16\linewidth]{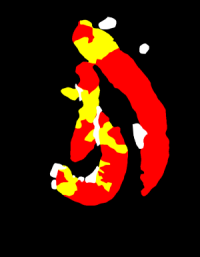}%
\label{GT_ihc}}
\hfill
\subfloat[]{\includegraphics[width=.16\linewidth]{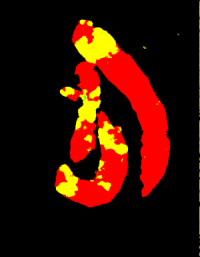}%
\label{cdunet_ihc}}
\hfill
\subfloat[]{\includegraphics[width=.16\linewidth]{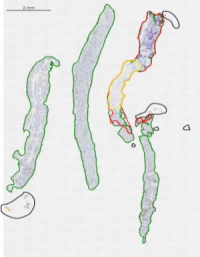}%
\label{orig_ihc_2}}
\hfill
\subfloat[]{\includegraphics[width=.16\linewidth]{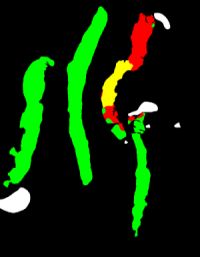}%
\label{GT_ihc_2}}
\hfill
\subfloat[]{\includegraphics[width=.16\linewidth]{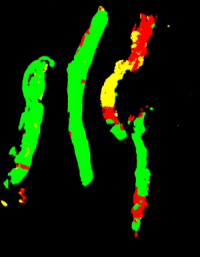}%
\label{cdunet_ihc_2}}

\caption{Examples of CD-UNET segmentation output of two images from the testing set. (a), (b) and (c) correspond to the original image, ground truth and corresponding segmentation result respectively of a CD163/CD68 slide. (d), (e) and (f) correspond to the same images for a CD8/KI67 slide. \label{exps_cdunet}}
\end{figure}

Figure \ref{exps_cdunet} shows some example results of the trained segmentation network on two differently stained slides.

\section{Tools for understanding the network}
Visualizing and understanding network decisions is especially important in the medical field where medical experts need to understand algorithm decisions in order to trust the results of automated analysis. For this purpose we visualize and highlight pixels in the image that were significant for the network's output (often termed attribution) as well as the features learned by the network (\citet{11, 30, 31, 32}). We also visualize outputs of specific layers.

\subsection{Color deconvolution segment visualization}
The visualization of filters and feature maps has been one of the most prominent tools used in deep learning in order to facilitate the understanding of the network decisions (\citet{33, 34}).
In order to visualize the effect of the color deconvolution segment on input images, we apply activation maximization to the filters of the first layer. Then we show some examples of the color deconvolution segment feature maps on different stainings.

\subsubsection{Activation maximization of the first layer filters}
\begin{figure}
\centering
\includegraphics[width=0.8\linewidth]{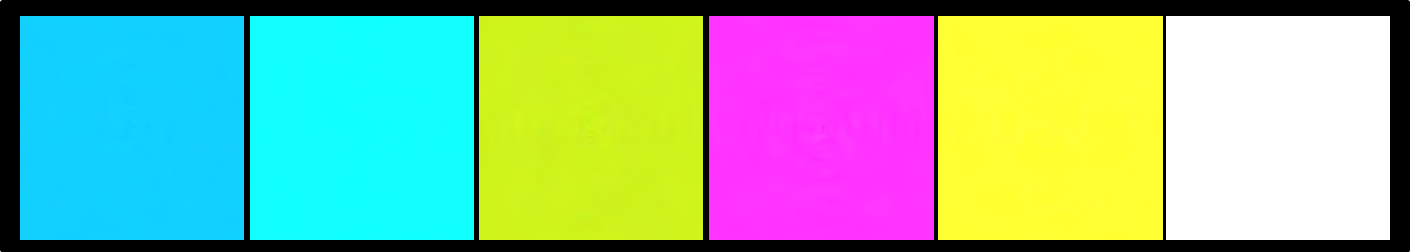}
\caption{Activation maximization of the filters of the first layer. The obtained images correspond to different stain colors. The white image corresponds to the background color.
}
\label{cdunet_filters_conv00}
\end{figure}

This approach allowed us to generate synthetic images that maximally activate the response of the first layer filters (\citet{40}). A noise image is inserted to the network, and several iterations of gradient ascent are run in order to modify the input image pixels to maximize the response of each of the filters. Figure \ref{cdunet_filters_conv00} shows the images we obtained following this approach. We notice that the resulted images correspond to stain colors from the training dataset.

\subsubsection{Color deconvolution segment output}
\begin{figure}
\centering
\subfloat[]{\includegraphics[width=.23\linewidth]{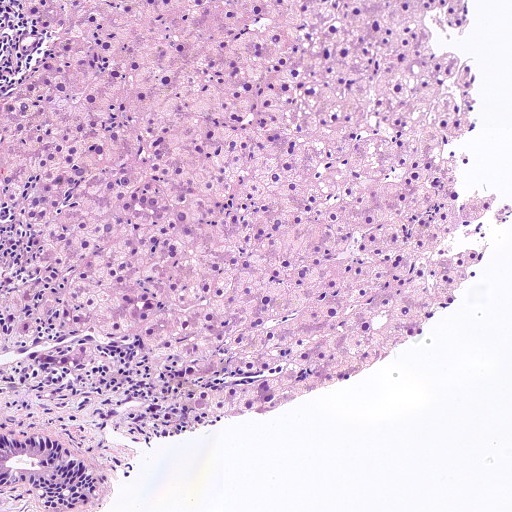}
\label{orig_1}}
\hfill
\subfloat[]{\includegraphics[width=.23\linewidth]{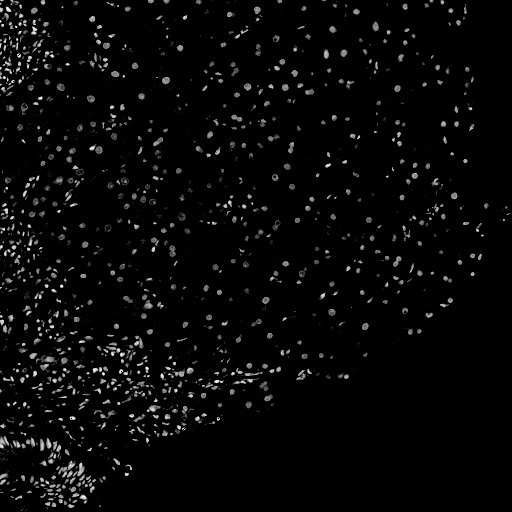}%
\label{out_1_1}}
\hfill
\subfloat[]{\includegraphics[width=.23\linewidth]{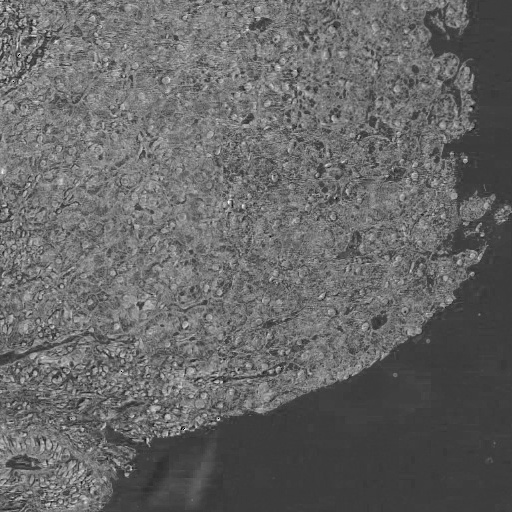}
\label{out_2_1}}
\hfill
\subfloat[]{\includegraphics[width=.23\linewidth]{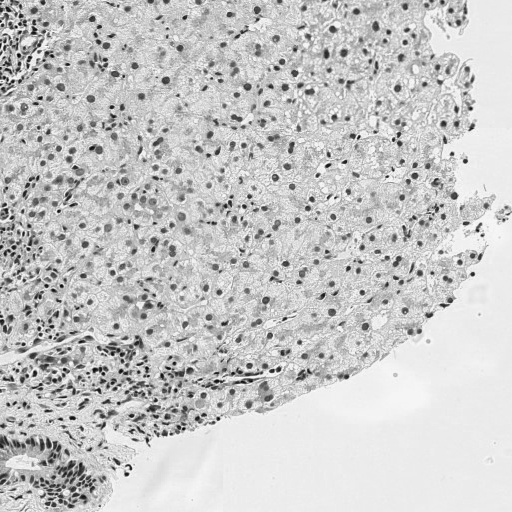}
\label{out_3_1}}

\caption{Example of the output of the color deconvolution segment on an H\&E image: (a) Original image (b) First output of the segment: corresponds to the hematoxylin channel (blue cells) (c) Second output of the segment: corresponds the the eosin channel (pink connective tissue) (d) Third output: corresponds to the background in this case. \label{cd_1}}
\end{figure}

In order to demonstrate the effect of the color deconvolution segment, we visualize its outputs using different stains (H\&E and IHC). Figure \ref{cd_1} shows an example of an input image and its corresponding outputs from the color deconvolution segment of the network.

%

\subsection{Feature visualization and pixel attribution}
\begin{figure}
\centering
\subfloat[Tumor]{\includegraphics[width=.3\linewidth]{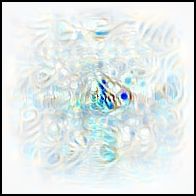}
\label{synt_1}}
\hfill
\subfloat[Tissue]{\includegraphics[width=.3\linewidth]{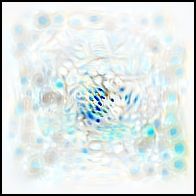}%
\label{synt_2}}
\hfill
\subfloat[Necrosis]{\includegraphics[width=.3\linewidth]{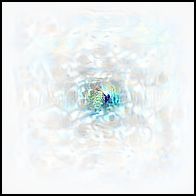}
\label{synt_3}}
\caption{(a), (b) and (c) correspond to synthetic images that maximize the scores for tumor, tissue and necrosis respectively. \label{synt_imgs}}
\end{figure}

For feature visualization we used the approach of \citet{14}. A noise image is inserted to the network, a specific pixel and category in the network output is set as the target, and several iterations of gradient ascent are run in order to modify the input image pixels to receive a high value in the target pixel. Using this we can create examples of input images, that cause a high activation at the target pixel for each of the categories (Figure \ref{synt_imgs}). An interesting observation is that the area of the pixels in the input image that affect an output pixel (the effective receptive field) is different for the three categories. The tissue category score is maximized when there are tissue cell nuclei far from the target pixel, implying that patterns of multiple tissue cell nuclei around the target pixel are used by the network as clues of tissue presence. The tumor and necrosis categories on the other hand seem to look for patterns of condensed large distorted cell nuclei around the target pixel. The synthetic image for tissue shows regular cell structures in the center but also far from the center meaning that patterns of cells around the target pixel were used by the network in order to make the decision.

\begin{figure}
\centering
\subfloat[Tumor]{\includegraphics[width=.32\linewidth]{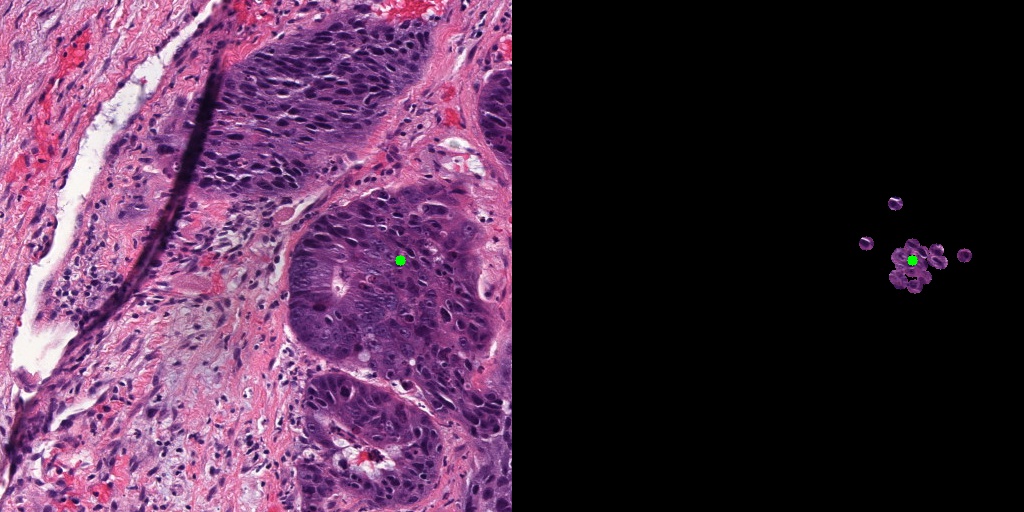}
\label{grad_1}}
\hfill
\subfloat[Tissue]{\includegraphics[width=.32\linewidth]{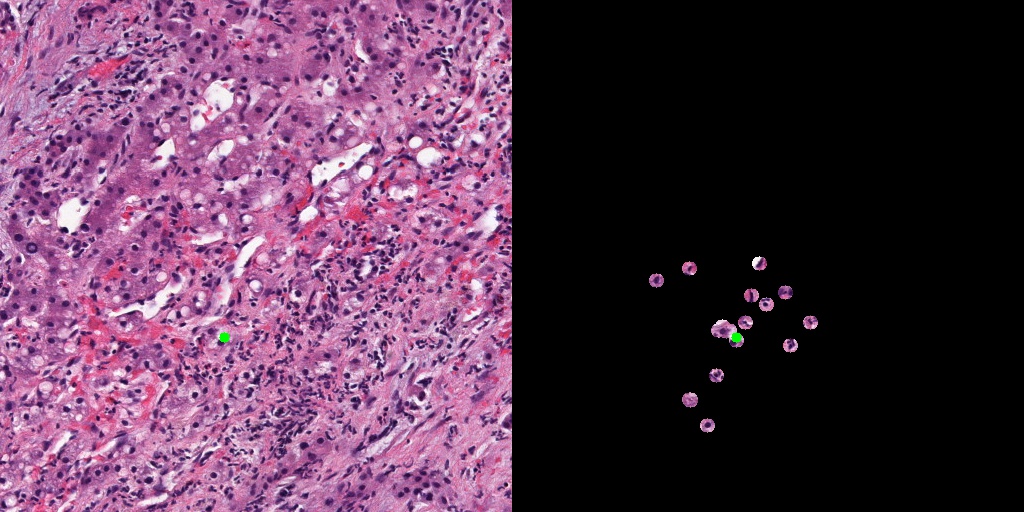}%
\label{grad_2}}
\hfill
\subfloat[Necrosis]{\includegraphics[width=.32\linewidth]{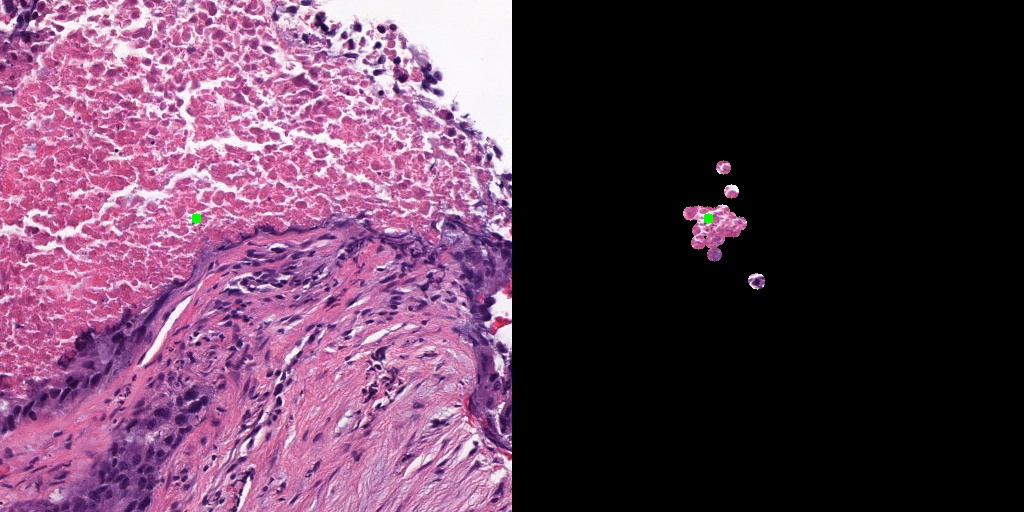}
\label{grad_3}}
\caption{(a), (b) and (c) correspond to SmoothGrad results for a tumor, a tissue and a necrosis pixel respectively. The target pixel is marked by the green circle. \label{grad_imgs}}
\end{figure}

For pixel attribution we used the SmoothGrad technique (\citet{15}), and averaged the gradients of the category score with respect to noisy input image pixels. The gradient image was passed to a ReLU gate, to keep only positive gradients. The gradient image is then thresholded, and input image pixels that had gradients below the threshold were masked out, to keep only pixels that were important for the network decision. A repeating theme in the visualizations was that for normal tissue and tumor, the gradients highlight respectively healthy looking and tumor cell nuclei in the target pixel surroundings, while ignoring other texture. Another important observation is that the effective receptive field was different between the different categories. The tissue category has a large effective receptive field compared to tumor and necrosis, which is in harmony with the interpretation of the synthetic images maximizing the scores of the classes. Figure \ref{grad_imgs} corresponds to the gradient images of the different categories.

\section{Discussion and conclusions}
Computerized segmentation of different tissue compartments has the advantage of being faster, less expensive, less laborious and more accurate and objective than manual segmentation traditionally performed by expert pathologists. Using expert annotated slides in order to teach an automatic segmentation model has the additional value of reusing expensive annotations and biopsy slide images to generate additional value. In the context of drug development, multiple IHC stains are used. To the best of our knowledge, this is the first work where multiple stains were simultaneously used in order to train a unified segmentation model that deals with multi stain histopathology images. Our experiments proved a higher difficulty in training the network for the more complex task of multistain image segmentation compared to the one stain scenario. Our interpretation was that the increased variability of the input image colors presented an additional complexity for the network and made the training process more difficult and erratic. 

Several state of the art methods used stain normalization as a pre-processing step (\citet{41, 42, 8}) in order to reduce color variability in the input space and improve segmentation and classification performance on different datasets. Other methods use color deconvolution as a pre-processing step. However, state of the art color deconvolution methods are highly dependent on the choice of a reference image, which is a very subjective and stain dependent task. Some methods suggested to include a stain separation layer as part of the network architecture (\citet{36}). However, they show that their method is highly dependent on filter initialization, which should follow stain basis vectors. Their architecture is also based on the assumption that only a single stain is used. In our case, different stains (H\&E and 8 IHCs) were used, and more stains are periodically added to our slide processing. In addition, defining stains reference vectors is a laborious and subjective task. We therefore present a method for generalizing tissue segmentation over multiple stainings by adding a color deconvolution segment to the segmentation network architecture. The parameters of this segment are optimized during the regular learning process in a "one-shot" training scheme. Adding the color deconvolution segment to UNET substantially improved the convergence smoothness and speed of the network when training on a multistain dataset. Generalization is also substantially improved, as can be seen by the network performance on the testing set. We theorize that this segment allows the network to deal with the variation of the input in the first color deconvolution layers and leaves the "rest" of the network with a much easier task to learn. The visualization of the outputs of the color deconvolution segment using different stainings as inputs shows the added layers actually learned to separate between different stain channels for the different stain types.

In order to enable understanding of the network architecture, we visualize synthetic images that maximize the scores of the different classes as well as different gradient images. This allowed us to see the effective receptive field that the network needs in order to make a decision for a specific pixel as well as to see what kinds of eigen-images make the network predict a specific label. 

We plan to continue investigating the effects of color deconvolution segments in other deep learning architectures and for additional digital pathology tasks, e.g. cell detection and classification. Additional variance capturing segments could be designed to help reducing dataset complexities from other sources, like different tumor and tissue types in order to facilitate inference when image sources are not known or when staining or imaging quality is suboptimal.

\subsubsection*{Acknowledgments}
The authors would like to thank and acknowledge the following people:
\begin{itemize}
\item Dr. Chen Sagiv for her detailed review of the text and valuable recommendations.
\item  Ben Levy for his valuable insights and code contributions.
\item Dr. Shadi Albarqouny  for his insights and fruitful discussions.
\item  Dr. Oliver Grimm for his trust and guidance. 
\item Dr. Fabien Gaire for his vision and continued support.
\end{itemize}


\bibliographystyle{unsrtnat}
\bibliography{segmentation}

%
%
%
%

\end{document}